\documentclass[12pt,letterpaper]{article}

\usepackage[
    letterpaper,
    top=1in,
    bottom=1in,
    left=1in,
    right=1in
]{geometry}
\usepackage{xcolor}
\usepackage{natbib}
\usepackage{hyperref}
\usepackage{dblfloatfix}
\usepackage{graphicx}
\usepackage{subcaption}
\usepackage{booktabs}
\usepackage{multirow}
\usepackage{enumitem}
\usepackage{amsmath,amssymb,amsthm}

\usepackage{booktabs}
\usepackage{algorithm}
\usepackage{algpseudocode}

\usepackage{graphicx}

\hypersetup{
    pdfauthor={Parvin Malekzadeh, Opher Baron, Dmitry Krass},
    pdftitle={A Unified Particle Filter LSTM for Data-Driven Process Simulation},
    pdfsubject={},
    pdfkeywords={data-driven process simulation, particle filter LSTM, event logs, emergency departments}
}
\usepackage{abstract}
\AtBeginDocument{
\setlength{\abovedisplayskip}{2pt}
\setlength{\belowdisplayskip}{2pt}
\setlength{\abovedisplayshortskip}{1pt}
\setlength{\belowdisplayshortskip}{1pt}
}

\title{\fontsize{16.9}{17}\selectfont
A Unified Particle Filter LSTM for Data-Driven Process Simulation}

\author{
{\small Parvin Malekzadeh$^{*}$ \quad Opher Baron \quad Dmitry Krass}\\[-0.03in]
{\footnotesize Rotman School of Management, University of Toronto, Toronto, ON, Canada}\\[-0.02in]
{\footnotesize $^{*}$Corresponding author: \href{mailto:p.malekzadeh@rotman.utoronto.ca}{p.malekzadeh@rotman.utoronto.ca}}
}

\date{}

\begin{document}

\maketitle
\begingroup
\renewcommand{\thefootnote}{}
\footnotetext{Accepted for presentation at the 2026 INFORMS Data Science Workshop.}
\endgroup

\vspace{-0.1in}
\vspace*{-0.2in}
\begin{abstract}
\vspace{-0.18in}
Data-driven process simulation aims to generate realistic case trajectories
from historical event logs without requiring an explicitly specified model of
the underlying dynamics. Deep sequence models can capture complex temporal
dependencies through next-activity probabilities and conditional time
distributions. However, event logs provide only a partial view of the
underlying process state, often recording activity completions without the
corresponding service-start times. Consequently, the same observed process
history may be consistent with multiple plausible latent process conditions,
whereas standard recurrent models compress each process prefix into a single
deterministic recurrent state.
We propose a Unified particle filter LSTM (Unified PF-LSTM) that maintains and sequentially
updates a weighted set of recurrent-state hypotheses. We summarize this
particle belief using its weighted mean and learned features based on the
moment-generating function. The resulting representation is used to predict a
categorical distribution over the next activity and conditional quantiles of
the current activity's sojourn time. The framework is trained end-to-end from
event-log data and evaluated on three real-world emergency department
datasets. The results show that the proposed framework consistently outperforms the considered data-driven baselines in reproducing routing, duration, and system-level behavior across all datasets, with particularly strong gains in settings where complex process dynamics are only partially reflected in the available event logs.
\end{abstract}

\vspace{-0.3in}
\section{Introduction} \label{sec:introduction}
\vspace{-0.1in}

Queueing and discrete-event simulation models are widely used to analyze congestion, evaluate operational policies, and support capacity-planning decisions. They also provide a foundation for digital twins and operational what-if analysis \citep{tao2018digital}. Constructing a high-fidelity simulator, however, generally requires a modeler to specify the system structure, arrival and service processes, routing logic, and resource interactions. This requires substantial domain knowledge and modeling expertise, particularly for systems with nonstationary, history-dependent, or partially observed dynamics. It also often requires collecting new customized data that may be difficult to maintain, and involves many subjective decisions by the modeler - two simulation experts are unlikely to arrive at the same model specification.

Modern information systems increasingly record operational processes as event logs containing event sequences, timestamps, case attributes, and contextual system information. Data-driven process simulation uses these records to learn generative models directly from observed trajectories, reducing the need to fully specify the underlying dynamics, and increasing modeling transparency and replicability. At each event, such a model must capture two related mechanisms: \emph{routing}, which determines where a case moves next, and \emph{sojourn time}, which determines how long it remains at the current activity. These outcomes depend on the case history and attributes, as well as evolving system conditions such as congestion.

Deep sequence models, including long short-term memory (LSTM) networks~\citep{LSTM}, are a natural approach to these tasks because they capture temporal dependencies in variable-length event histories \citep{camargo2019learning,gunnarsson2023direct}.
 Standard deep sequence models, however, typically compress each observed process prefix (i.e., the case’s event history up to the current point) into a single deterministic recurrent state, a learned numerical representation of the observed history.

Event-log data provide only a partial view of the underlying process state. In particular, many real-world event logs contain only one timestamp per activity, typically its completion time, while the service-start time is unavailable \citep{fracca2022estimating,suriadi2015event}. Consequently, the elapsed time between consecutive activity completions conflates waiting and service time, so the same observed interval may correspond to different underlying operational conditions. More generally, relevant case characteristics may be unavailable, system measurements may be noisy or delayed, and factors such as effective resource availability or unrecorded workload may not be observed. The same observed history may therefore be compatible with several plausible latent process states. A single recurrent state represents only one interpretation of the observed history and does not explicitly preserve this latent-state ambiguity.
\\
This issue is particularly important during recursive generation, where sampled activities and sojourn times become inputs to subsequent predictions. An inaccurate recurrent representation at one step can influence the remaining trajectory, allowing prediction errors to propagate and compound.

\vspace{.1in}
\noindent\textbf{Contributions.}
To address this limitation, we build on particle filter LSTM (PF-LSTMs) \citep{ma2020pfrnn}. Rather than maintaining a single recurrent state, a PF-LSTM maintains a weighted particle approximation of the belief over recurrent states. It updates this belief through an importance-weighted particle-filter procedure implemented as a differentiable computational graph. The resulting representation preserves multiple plausible recurrent interpretations of the observed history and sequentially updates their relative importance.

Using the complete particle set directly for prediction is difficult, while its weighted mean alone may discard information about the shape of the belief. We therefore augment the weighted mean with learned features based on the moment-generating function (MGF) \citep{bulmer1979principles}. These features are permutation-invariant, computationally efficient, statistically sufficient for many queuing representations, and easy to optimize, especially when the particle set is large \citep{johnson2019statistics}.

The resulting fixed-dimensional belief representation is shared by two prediction heads: a routing head that produces a categorical distribution over the next activity and a timing head that estimates conditional quantiles of the current activity's sojourn time. 
\\
Our main contributions are:
\vspace{-0.1in}
\begin{enumerate}[leftmargin=*, labelsep=0.5em]
\item We adapt a PF-LSTM with MGF-based belief features to data-driven process modeling, representing latent-state uncertainty induced by partial observability while also modeling intrinsic variability in routing and sojourn-time outcomes. Although instantiated using an LSTM, the underlying framework is applicable to other sequential architectures.
\vspace{-0.1in}
\item We demonstrate the efficacy of the framework using data from three emergency departments (EDs) with over 120,000 patient visits and 1,200,000 station visits through routing performance, duration calibration, and system-level fidelity. We observe that while our framework requires longer runtime, it leads to substantially higher accuracy, particularly for complex processes whose event logs provide limited information about the underlying system state.
\end{enumerate}
%
\noindent\textbf{Related Work.}
Work on data-driven process modeling includes waiting-time prediction, patient-flow forecasting, queue-performance estimation, and generative modeling of queueing systems \citep{ang2016waiting,sharafat2021patient,baron2024queue,ojeda2021generative}.

Machine learning methods such as random forests and gradient boosting provide flexible, nonparametric models of process outcomes. Quantile-based tree ensembles can additionally estimate conditional outcome distributions and prediction intervals \citep{mehdiyev2025quantifying}. These methods, however, typically rely on fixed-dimensional representations and do not directly capture dependencies across variable-length event histories.

Deep sequence models address this limitation and include recurrent architectures such as LSTMs \citep{tax2017predictive,camargo2019learning,camargo2021discovering, gunnarsson2023direct} and Transformer-based models \citep{mittal2025data}. Probabilistic predictions are particularly important for simulation: next-activity probabilities represent variability in routing, while conditional sojourn time distributions represent variability in activity durations~\citep{mittal2025data,  mehdiyev2025quantifying}. However, these models typically do not represent latent-state uncertainty from incomplete event-log observations. Our work addresses this gap through a particle-filter mechanism that maintains multiple weighted recurrent-state hypotheses.


\vspace{-0.2in}
\section{Problem Formulation}
\label{sec:problem}
\vspace{-0.1in}
Operational data are typically available as event tables that record the sequence of activities visited by each case and the corresponding timestamps, together with static case attributes such as age and gender. We represent the activity trace of a case as
\(
\sigma=\big\langle (A_1,T_1),\ldots,(A_N,T_N)\big\rangle, 
\)
where \(A_k\in\mathcal A\) denotes the activity visited at step \(k\), and \(T_k\geq 0\) is the time spent until transition to the next event.

At each event, we augment the recorded process history with a vector
$\mathbf{x}_k$ of dynamic features describing the current case and system
conditions. These features may include elapsed process time, the number of cases
in the system, activity-level census, and other congestion measures. Let
$\mathbf{z}$ denote the static case attributes. We define the observation (input) at
step $k$ as
\(
\mathbf{o}_k
=
\left(A_k,T_{k-1},\mathbf{x}_k\right),
\)
where $T_{k-1}$ is omitted for the first event. The information available through
step $k$ is then
\begin{equation}
\mathcal{H}_k
=
\left(\mathbf{o}_1,\ldots,\mathbf{o}_k,\mathbf{z}\right)
=
\left(A_{1:k},T_{1:k-1},\mathbf{x}_{1:k},\mathbf{z}\right).
\label{Eq:observation_history}
\end{equation}

Given $\mathcal{H}_k$ at each step $k \in \{1, 2, ..., N \}$, the objective is to estimate the conditional distribution of the next activity, 
\begin{equation} 
\Pr\!\left(A_{k+1}=a \mid \mathcal{H}_k\right), \qquad a \in \mathcal{A}, \quad \text{with } A_{N+1}=\texttt{END},
\end{equation} 
and the distribution of the sojourn time at the activity entered at this event. We represent the latter using $N_q$ quantiles, 
\begin{equation}
Q_{\tau_n}\!\left(T_k \mid \mathcal{H}_k\right), \qquad \tau_n=\frac{n}{N_q+1}, \qquad n=1,\ldots,N_q, 
\end{equation}
where $\tau_n \in (0,1)$ denotes the corresponding quantile level. A quantile-based representation avoids imposing a particular parametric family on the distribution of event durations, which may be skewed, heavy-tailed, or heteroscedastic.

\vspace{-0.2in}
\section{Methodology}
\label{sec:methodology}
\vspace{-0.1in}
Event-log data provide only a partial view of the underlying process state. Although the dynamic feature vector $\mathbf{x}_k$ captures observable system conditions, some factors that influence case routing and activity sojourn times may be unavailable, measured with error, or recorded with delay. For example, because only activity-completion timestamps are available, the elapsed time used as the activity sojourn time does not distinguish waiting from service time. Resource availability, unrecorded workload, and latent case characteristics may also be unavailable or measured imperfectly. Consequently, the same observed history $\mathcal{H}_k$ may be consistent with multiple plausible underlying process conditions.

To represent the resulting latent-state uncertainty, we use a PF-LSTM architecture; see \citet{ma2020pfrnn} for further details. We first provide a brief overview of the PF-LSTM and then present our unified framework, which integrates the particle-belief representation with routing and timing prediction, end-to-end training, and recursive simulation.

\vspace{-0.15in}
\subsection{Particle Filter LSTM (PF-LSTM)}
\label{sec:pflstm}
\vspace{-0.1in}
A standard LSTM maps $\mathcal{H}_k$ to a single recurrent state. In contrast, a PF-LSTM \citet{ma2020pfrnn} maintains a weighted set of recurrent-state hypotheses. Each particle is a learned neural representation of a plausible latent process condition consistent with the observed history, rather than a direct estimate of a physical process state. Collectively, the particles form a learned belief representation for the routing and timing prediction tasks.

Before processing the observation $\mathbf{o}_k = \left(A_k,T_{k-1},\mathbf{x}_k\right)$, the particle belief is
\begin{equation}
\mathcal{B}_{k-1}
=
\left\{
\left(
\mathbf{h}_{k-1}^{(i)},
\mathbf{c}_{k-1}^{(i)},
w_{k-1}^{(i)}
\right)
\right\}_{i=1}^{P},
\end{equation}
where $\mathbf{h}_{k-1}^{(i)}$ is the hidden state of particle $i$, $\mathbf{c}_{k-1}^{(i)}$ is its internal LSTM cell state, and $w_{k-1}^{(i)}$ is its normalized weight. The hidden state is exposed to the downstream belief representation and prediction heads, whereas the cell state serves as the particle's internal recurrent memory. The pair $\left(\mathbf{h}_{k-1}^{(i)},\mathbf{c}_{k-1}^{(i)}\right)$ therefore forms the recurrent state of particle $i$, and both components are propagated and resampled together. Here, $P$ denotes the number of particles.

\textbf{1. Stochastic particle transition.}
Each particle is propagated through a shared stochastic PF-LSTM transition:
\begin{equation}
\left(
\widetilde{\mathbf{h}}_k^{(i)},
\widetilde{\mathbf{c}}_k^{(i)}
\right)
=
g_{\theta}^{\mathrm{transit}}
\left(
\mathbf{h}_{k-1}^{(i)},
\mathbf{c}_{k-1}^{(i)},
\mathbf{o}_k,
\boldsymbol{\epsilon}_k^{(i)}
\right),
\end{equation}
where $g_{\theta}^{\mathrm{transit}}$ is a learnable PF-LSTM transition mapping
parameterized by $\theta$, and
$\boldsymbol{\epsilon}_k^{(i)}$ is independently sampled Gaussian noise whose
distribution is parameterized using the previous particle state and current
observation. The stochastic term allows the particles to represent different
recurrent-state hypotheses and helps preserve particle diversity.

\textbf{2. Weight update.}
Each propagated particle receives a positive compatibility score
\begin{equation}
\ell_k^{(i)}
=
g_{\phi}^{\mathrm{weight}}
\left(
\mathbf{o}_k,
\widetilde{\mathbf{h}}_k^{(i)}
\right),
\qquad
\ell_k^{(i)}>0,
\end{equation}
where $g_{\phi}^{\mathrm{weight}}$ is a learnable scoring function parameterized by $\phi$. Its
normalized weight is then updated as
\(
\widetilde{w}_k^{(i)}
=
\frac{
w_{k-1}^{(i)}\ell_k^{(i)}
}{
\sum_{j=1}^{P}
w_{k-1}^{(j)}\ell_k^{(j)}
}.
\)


\textbf{3. Soft resampling.}
Finally, ancestor particles are sampled from the soft-resampling distribution
$q_k(i)=\alpha\widetilde{w}_k^{(i)}+(1-\alpha)/P$, where $\alpha\in[0,1]$ balances weight-based and uniform sampling.
An importance-weight correction is then applied to account for the modified sampling distribution, yielding the updated particle belief $\mathcal{B}_k$.

\vspace{-0.2in}
\subsection{The proposed Unified PF-LSTM}
\label{sec:unified_pflstm}
\begin{figure}[!bp]
\vspace{-0.2in}
    \centering
    \includegraphics[width=\textwidth]{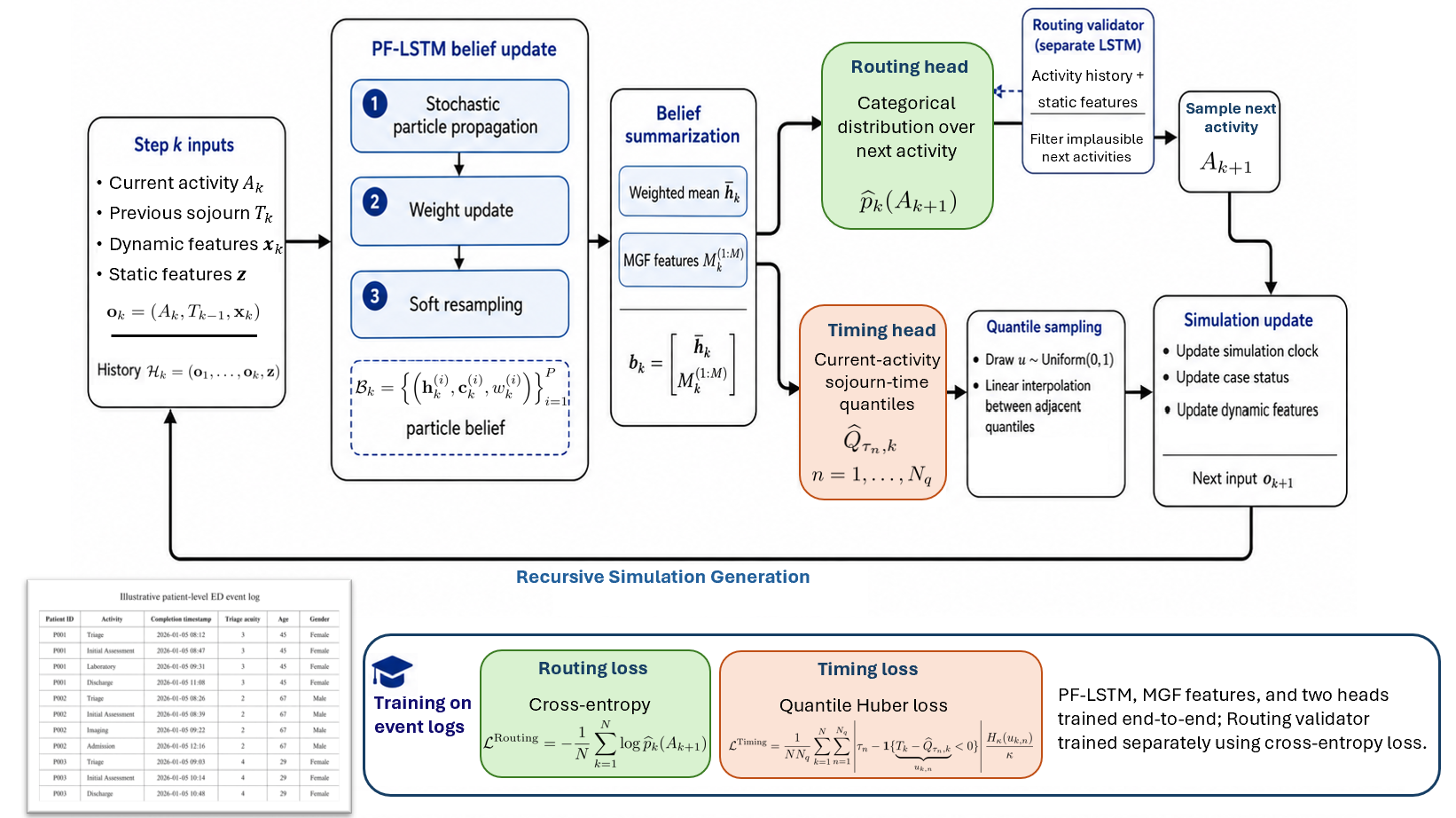}
    \vspace{-.25in}
    \caption{Overview of the proposed unified PF-LSTM framework. }
    \label{fig:unified_pflstm}
    \vspace{-0.2in}
\end{figure}

\vspace{-0.1in}
Figure~\ref{fig:unified_pflstm} summarizes our proposed Unified PF-LSTM
architecture. This PF-LSTM first processes the observation sequence using the
particle-filter update described in Section~\ref{sec:pflstm}. At each step $k$,
this produces the weighted particle belief
\(
\mathcal{B}_k
=
\left\{
\left(
\mathbf{h}_k^{(i)},
\mathbf{c}_k^{(i)},
w_k^{(i)}
\right)
\right\}_{i=1}^{P}.
\)
The particle belief is then summarized by a fixed-dimensional representation
for prediction.

We compute the weighted mean of the particle hidden states,
\(
\overline{\mathbf{h}}_k
=
\sum_{i=1}^{P}
w_k^{(i)}\mathbf{h}_k^{(i)}.
\)
To retain information about the belief beyond its mean, we augment
$\overline{\mathbf{h}}_k$ with $M$ features based on the MGF. In statistics, the MGF provides an alternative characterization of a random variable's distribution and can be used to generate its moments~\citep{johnson2019statistics}. Treating the weighted particle hidden states as an empirical distribution, its MGF is
\[
\mathcal{M}_k(\mathbf{v}) = \sum_{i=1}^{P} w_k^{(i)} \exp\!\left( \mathbf{v}^{\top}\mathbf{h}_k^{(i)} \right), 
\]
where $\mathbf{v}\in\mathbb{R}^{d_h}$ is the argument at which the MGF is evaluated.
We evaluate the MGF at $M$ trainable vectors $\mathbf{v}_1,\ldots,\mathbf{v}_M$. Each $\mathbf{v}_m\in\mathbb{R}^{d_h}$ is a parameter that defines a linear projection of the particle state. The corresponding feature is
\( M_k^{(m)} = \mathcal{M}_k(\mathbf{v}_m). \)
In the neural-network implementation, the vectors $\mathbf{v}_1,\ldots,\mathbf{v}_M$ form the rows of a trainable linear layer applied to each particle state, followed by an elementwise exponential activation and a weighted aggregation across particles.
The resulting belief representation is
\(
\mathbf{b}_k
=
\left[
\overline{\mathbf{h}}_k;\,
M_k^{(1:M)}
\right].
\)
%

This belief representation is shared by two task-specific fully connected prediction heads. Let $g_{\psi}^{\mathrm{route}}$ and $g_{\zeta}^{\mathrm{timing}}$ denote the routing and timing networks, with parameters $\psi$ and $\zeta$, respectively. The routing network maps the belief representation $\mathbf{b}_k$ to a categorical distribution over the next activity, while the timing network maps it to the $N_q$ conditional quantiles of the current activity's sojourn time.

In addition, recursive simulation can amplify errors caused by extreme sampled
sojourn times. Each sampled sojourn time affects the simulated
system state and, consequently, the dynamic feature vector constructed at the next step. Unusually large or small values may create
system conditions rarely observed in the training data. Under such conditions,
the main routing network may assign nonzero probability to implausible next
activities.

To improve robustness, we use a separately trained LSTM routing validator. The
validator relies only on the activity history and static case attributes and
does not use the dynamic system features affected by the sampled sojourn times.
At each step, it identifies plausible next activities, which are
used to filter the categorical distribution produced by the main routing
network before the next activity is sampled. The remaining probabilities are
then renormalized. Thus, the validator does not replace the main routing
network; rather, it acts as a safeguard against implausible transitions caused
by outlier-induced system conditions during recursive generation.

\vspace{.1in}
\noindent\textbf{Training.}
Let $\widehat{p}_k(a)$ denote the predicted routing probability for activity
$a$, and let $\widehat{Q}_{\tau_n,k}$ denote the predicted
$\tau_n$-quantile of $T_k$. The routing head is trained using categorical
cross-entropy loss,
\begin{equation}
\mathcal{L}^{\mathrm{Routing}}
=
-\frac{1}{N} \sum_{k=1}^{N}
\log \widehat{p}_k(A_{k+1}).
\end{equation}

For the timing head, define the residual
$u_{k,n}=T_k-\widehat{Q}_{\tau_n,k}$. We use the quantile Huber loss
\citep{huber1992robust,dabney2018distributional},
\begin{equation}
\mathcal{L}^{\mathrm{Timing}}
=
\frac{1}{N N_q} \sum_{k=1}^{N}\sum_{n=1}^{N_q}
\left|
\tau_n-\mathbf{1}\{u_{k,n}<0\}
\right|
\frac{H_{\kappa}(u_{k,n})}{\kappa},
\end{equation}
where $\kappa>0$ is the Huber threshold and
\begin{equation}
H_{\kappa}(u)
=
\begin{cases}
\frac{1}{2}u^2, & |u|\leq\kappa,\\[1mm]
\kappa\left(|u|-\frac{1}{2}\kappa\right), & |u|>\kappa.
\end{cases}
\end{equation}
The overall training objective is
\(
\mathcal{L}
=
\mathcal{L}^{\mathrm{Routing}}
+
\mathcal{L}^{\mathrm{Timing}}.
\)

Let
\(
\mathbf{V} = \left[\mathbf{v}_1,\ldots,\mathbf{v}_M\right]^\top \! \in \!\mathbb{R}^{M\times d_h}
\)
denote the trainable MGF matrix, and let
\(
\Omega
=
\{\theta,\phi,\mathbf{V},\psi,\zeta\}
\)
collect the trainable parameters of the Unified PF-LSTM. Define \( \widehat{\mathbf{p}}_k = \left(\widehat{p}_k(a)\right)_{a\in\mathcal{A}} \) and \( \widehat{\mathbf{Q}}_k \!= \! \left( \widehat{Q}_{\tau_1,k}, \ldots, \widehat{Q}_{\tau_{N_q},k} \right)^\top. \)
\\
Writing $\mathcal{L}^{\mathrm{Routing}}_k$ and
$\mathcal{L}^{\mathrm{Timing}}_k$ for the corresponding per-step losses, the
gradient passed from the two prediction heads to the shared belief
representation is
\begin{equation}
\boldsymbol{\delta}_k^{b}
=
\left(
\frac{\partial \widehat{\mathbf{p}}_k}
     {\partial \mathbf{b}_k}
\right)^{\!\top}
\nabla_{\widehat{\mathbf{p}}_k}
\mathcal{L}^{\mathrm{Routing}}_k
+
\left(
\frac{\partial \widehat{\mathbf{Q}}_k}
     {\partial \mathbf{b}_k}
\right)^{\!\top}
\nabla_{\widehat{\mathbf{Q}}_k}
\mathcal{L}^{\mathrm{Timing}}_k.
\end{equation}
Therefore, for any shared parameter
$\omega\in\{\theta,\phi,\mathbf{V}\}$,
\(
\nabla_{\omega}\mathcal{L}
=
\sum_{k=1}^{N}
\left(
\frac{\mathrm{d}\mathbf{b}_k}
     {\mathrm{d}\omega}
\right)^{\!\top}
\boldsymbol{\delta}_k^{b}.
\)
The total derivative accounts for the dependence of the particle recurrent
states and weights on preceding steps and is computed using backpropagation
through time. Because the MGF features are included in $\mathbf{b}_k$,
gradients from both losses propagate through these features to update
$\mathbf{V}$. The routing-head parameters $\psi$ receive gradients only from
$\mathcal{L}^{\mathrm{Routing}}$, whereas the timing-head parameters $\zeta$
receive gradients only from $\mathcal{L}^{\mathrm{Timing}}$. Thus, the
particle-transition network, particle-weight network, MGF matrix, and two
prediction heads are trained end to end.

The routing validator is trained separately using a next-activity
cross-entropy objective, and its parameters are not included in $\Omega$. 

\vspace{.1in}
\noindent\textbf{Simulation.}
Once trained, the Unified PF-LSTM generates case trajectories
autoregressively. At step $k$, the routing head produces a categorical
distribution $\widehat{p}_k(a)$ over the possible next activities. The
separately trained routing validator then evaluates the plausibility of each
candidate activity using only the activity history and static case attributes.
Activities whose validator probabilities fall below a predefined threshold are
removed from the candidate set. The probabilities assigned by the main routing
head to the remaining activities are then renormalized, and the next activity
$A_{k+1}$ is sampled from the resulting distribution.

If the validator removes all candidates, the activity with the highest
validator probability is retained. Thus, the validator restricts the set of
admissible next activities without replacing the routing distribution learned
by the Unified PF-LSTM.

To sample the duration of the current activity, we interpret the 
conditional quantiles as a piecewise-linear approximation of the inverse
conditional distribution. We draw
$u\sim\operatorname{Uniform}(0,1)$. For adjacent quantile levels satisfying
$\tau_n\leq u\leq\tau_{n+1}$, the sampled duration is
\begin{equation}
\widetilde{T}_k
=
\widehat{Q}_{\tau_n,k}
+
\frac{u-\tau_n}{\tau_{n+1}-\tau_n}
\left(
\widehat{Q}_{\tau_{n+1},k}
-
\widehat{Q}_{\tau_n,k}
\right).
\end{equation}
For $u<\tau_1$, we use $\widehat{Q}_{\tau_1,k}$, and for
$u>\tau_{N_q}$, we use $\widehat{Q}_{\tau_{N_q},k}$.

Upon arrival at activity $A_k$, the model samples the current activity's
sojourn time $\widetilde{T}_k$ and the next activity $A_{k+1}$. The transition to $A_{k+1}$ is scheduled for the current simulation time plus
$\widetilde{T}_k$. When the transition occurs, the case status,
activity-level census, and other dynamic system features are updated,
producing $\mathbf{x}_{k+1}$. The next model input is
\(
\mathbf{o}_{k+1}
=
\left(
A_{k+1},
\widetilde{T}_k,
\mathbf{x}_{k+1}
\right).
\)
The model parameters remain fixed during generation, while the particle recurrent states
and weights are updated at each step as the generated trajectory evolves. This
procedure continues until the terminal activity $\texttt{END}$ is generated.

\vspace{-0.2in}
\section{Experimental Evaluation}
\label{sec:experiments}
\vspace{-0.1in}

In this section, we empirically evaluate the capabilities of the proposed Unified PF-LSTM framework for learning both routing and activity sojourn-time
distributions.

\begin{figure*}[!tbp]
    \vspace{-.2in}
    \centering
    \captionsetup[subfigure]{skip=1pt}
    \captionsetup{skip=2pt}

    \begin{subfigure}[t]{0.145\textwidth}
        \centering
        \includegraphics[width=\linewidth]
        {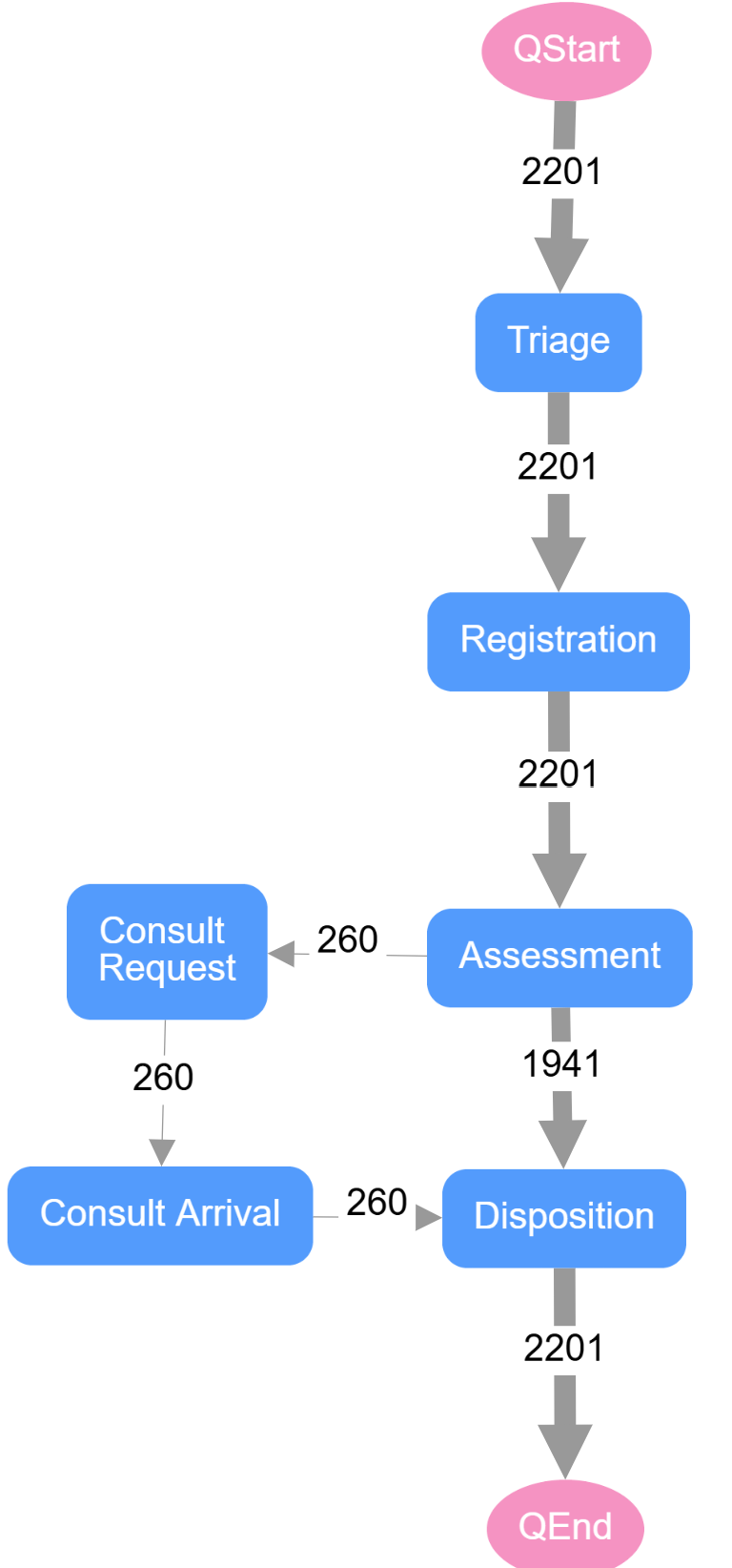}
        \caption{ED A.}
        \label{fig:process_ed_a}
    \end{subfigure}
    \hfill
    \begin{subfigure}[t]{0.35\textwidth}
        \centering
        \includegraphics[width=\linewidth]
        {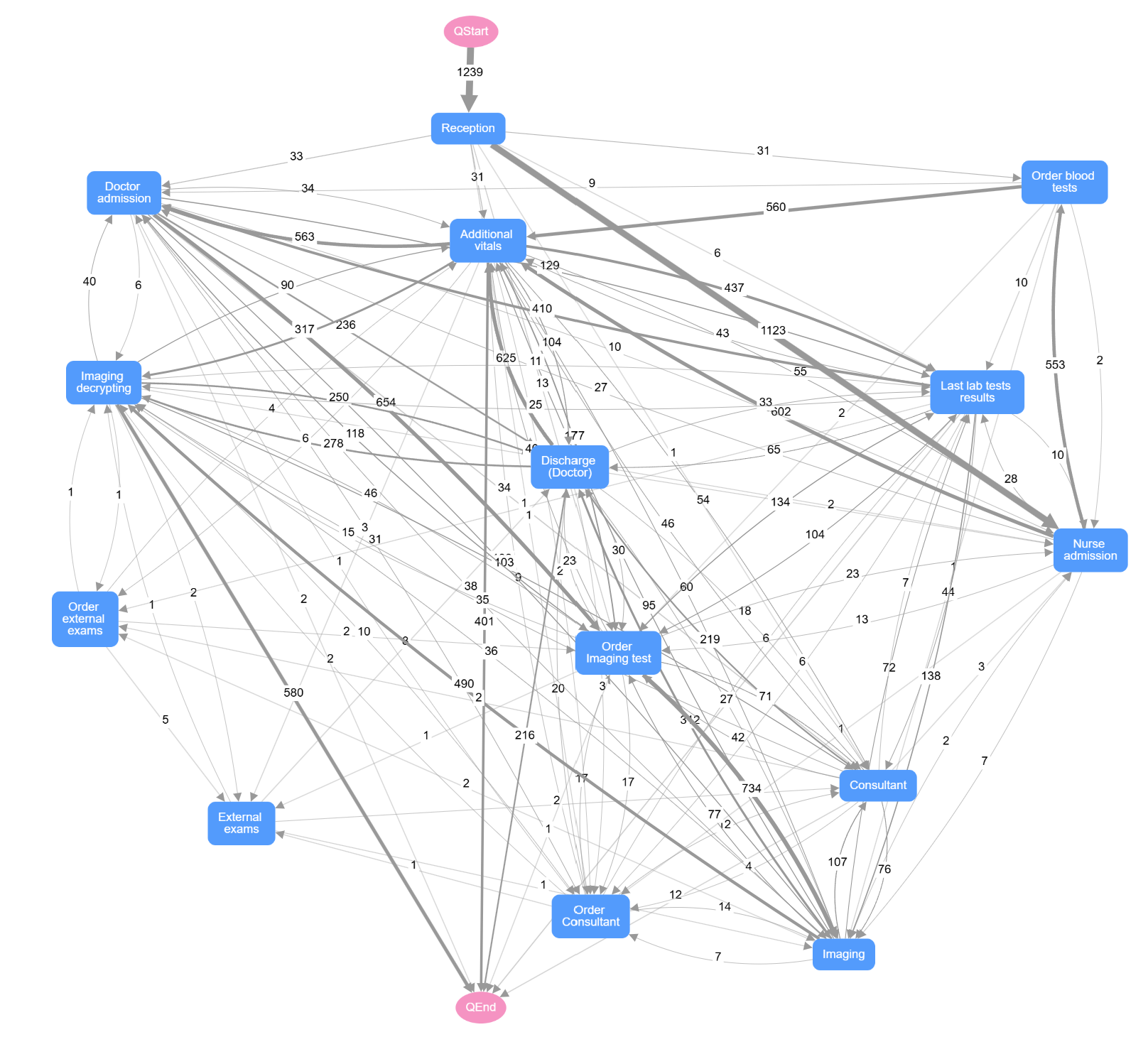}
        \caption{ED B.}
        \label{fig:process_ed_b}
    \end{subfigure}
    \hfill
    \begin{subfigure}[t]{0.48\textwidth}
        \centering
        \includegraphics[width=\linewidth]
        {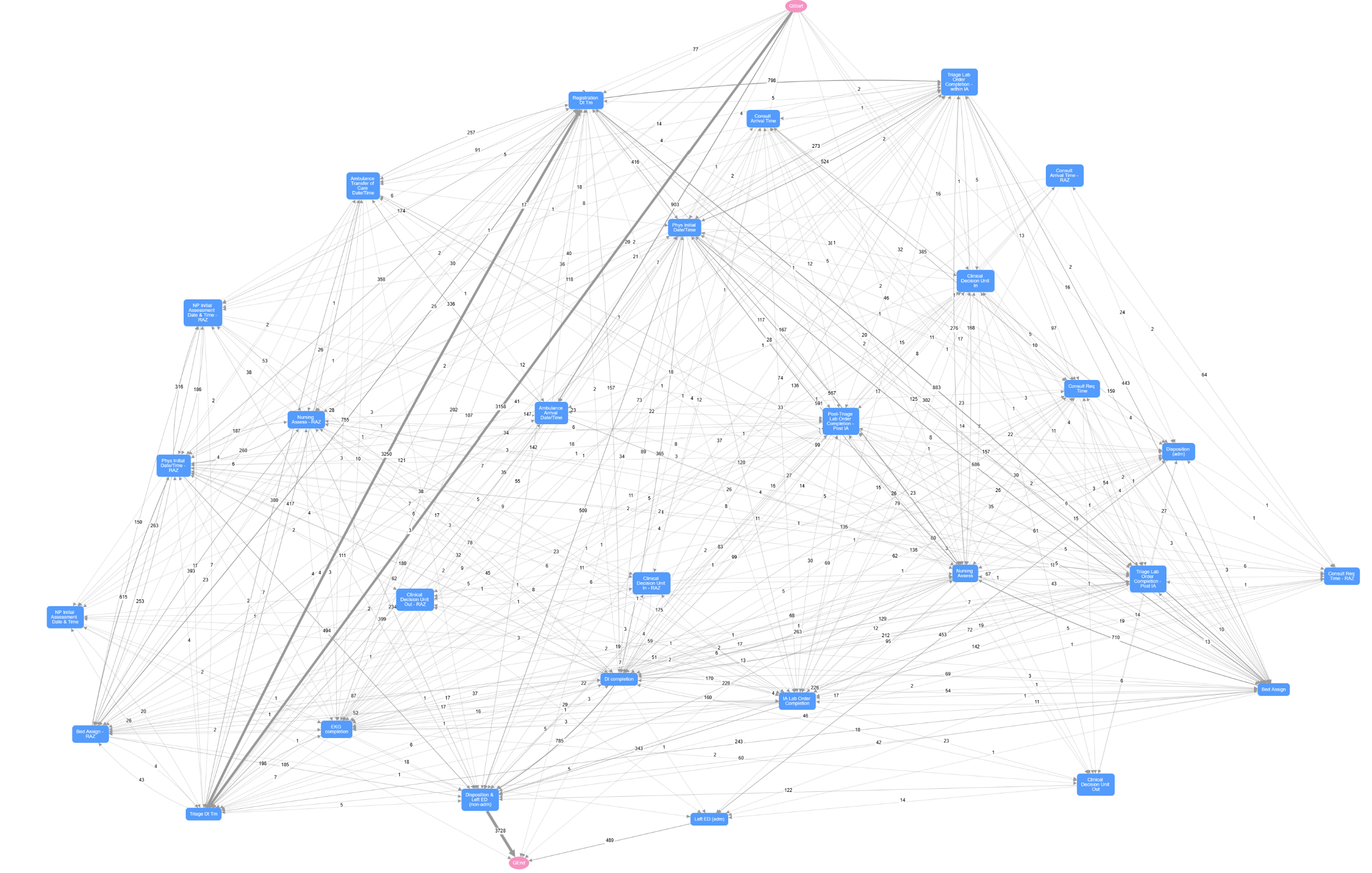}
        \caption{ED C.}
        \label{fig:process_ed_c}
    \end{subfigure}

    \caption{Process maps for the three ED datasets. Pink circular nodes
    indicate the start and end of a patient trajectory, while blue rectangular
    nodes represent ED activities or stations. Directed edges represent
    transitions observed in the corresponding event log.}
    \label{fig:ED_process_maps}
    \vspace{-.2in}
\end{figure*}

\vspace{0.1in}
\noindent\textbf{Dataset overview.}
We use data from three EDs, referred to as ED A, ED B, and ED C.\footnote{The
hospital names are withheld for confidentiality.} Figure~\ref{fig:ED_process_maps} depicts their process maps: ED A has a relatively simple process structure, ED B has a moderately complex structure, and ED C has the most complex structure.

Each dataset consists of patient-level event logs that record the sequence of
ED activities or stations visited by each patient and the completion timestamp
of each activity. As is common in many applications~\citep{fracca2022estimating,suriadi2015event}, these data do not include timestamps indicating when service begins; only the completion times of the activities are observed. For activity $A_k$, the completion time of $A_{k-1}$ is treated as the arrival time to $A_k$, and the sojourn time at $A_k$ is calculated as the difference between the two consecutive completion timestamps.
For the first activity, the sojourn time is measured from the recorded ED
arrival time to the completion time of that activity. The datasets also contain
static patient attributes, such as triage acuity, age, and gender, subject to
their availability at each hospital. ED A contains 11,652 patient trajectories,
ED B contains 30,876, and ED C contains 82,435.

Upon a patient's arrival at activity $A_k$, we construct the dynamic feature
vector $\mathbf{x}_k$ using information available at that time. These
features include the activity-level census, the total number of patients in the ED, and measures of residual workload at the relevant
activities.

\vspace{0.1in}
\noindent\textbf{Baselines.}
We compare the proposed Unified PF-LSTM with a hybrid framework that combines
a fifth-order Markov model for routing with a random survival forest (RSF) for
sojourn-time prediction. The Markov model captures dependence on the five most
recent activities, while RSF has been used as a state-of-the-art nonparametric
method for estimating conditional quantiles of sojourn times
\citep{mehdiyev2025quantifying}. This baseline models routing and timing
separately and provides a strong non-neural benchmark. We also consider an
otherwise comparable LSTM-based model with a routing validator but no particle filtering to isolate the contribution of the particle-based belief representation.

\vspace{0.1in}
\noindent\textbf{Model training setup.}
For each dataset, patient encounters are partitioned chronologically, with the
earliest 80\% used for training and the most recent 20\% reserved for testing.
All events belonging to the same encounter are retained within a single
partition. A validation subset of the training data is used for hyperparameter
selection.

Across all datasets, the Unified PF-LSTM uses $P=20$ particles, $M=8$ MGF features, and a soft-resampling coefficient of $\alpha=0.5$. We examine the sensitivity of model performance to $P$ and $M$ in Section~\ref{sec:sensitivity}. Table~\ref{tab:hyperparameters} reports the architectural and training hyperparameters shared by the Unified PF-LSTM and the LSTM baseline.  
\begin{table*}[!hbp]
\vspace{-0.1in}
\centering
\captionsetup{skip=2pt}
\caption{Hyperparameters shared by the Unified PF-LSTM and LSTM baseline.}
\label{tab:hyperparameters}
\setlength{\tabcolsep}{4pt}
\renewcommand{\arraystretch}{1.0}
\resizebox{\textwidth}{!}{
\begin{tabular}{@{}lcccccc@{}}
\toprule
\textbf{Parameter}
& \# Recurrent layers
& Hidden-state dimension ($d_h$)
& Embedding dimension
& Dropout rate
& \# Quantiles ($N_q$)
& Huber threshold ($\kappa$)
\\
\midrule
\textbf{Value}
& 2
& 128
& 64
& 0.1
& 20
& 1
\\
\midrule
\textbf{Parameter}
& Routing-validator threshold
& Optimizer
& Learning rate 
& Weight decay 
& Batch size 
& Maximum epochs
\\
\midrule
\textbf{Value}
& 0.1
& Adam
& $10^{-3}$
& 0.01
& 64
& 30
\\
\bottomrule
\end{tabular}
}
\vspace{-0.15in}
\end{table*}

\vspace{0.1in}
\noindent\textbf{Model evaluation.}
Because the competing models use different training objectives and output
representations, we evaluate their end-to-end simulation performance. For each
held-out test case, a model recursively samples the current activity's sojourn
time and the next activity until the generated patient trajectory reaches the
terminal activity. Each experiment is run independently 10 times, and the reported results are
averaged across runs.

We compare the observed and simulated outcomes using time to provider initial assessment
(TPIA) and total length of stay (LOS), two key performance indicators of ED
operations~\citep{cihi2018waittimes}. 

\begin{figure}[h]
    \vspace{-0.1in}
    \centering
    \captionsetup[subfigure]{skip=1pt}
    \captionsetup{skip=2pt}

    \begin{subfigure}[t]{0.3\textwidth}
        \centering
        \includegraphics[width=\linewidth]
        {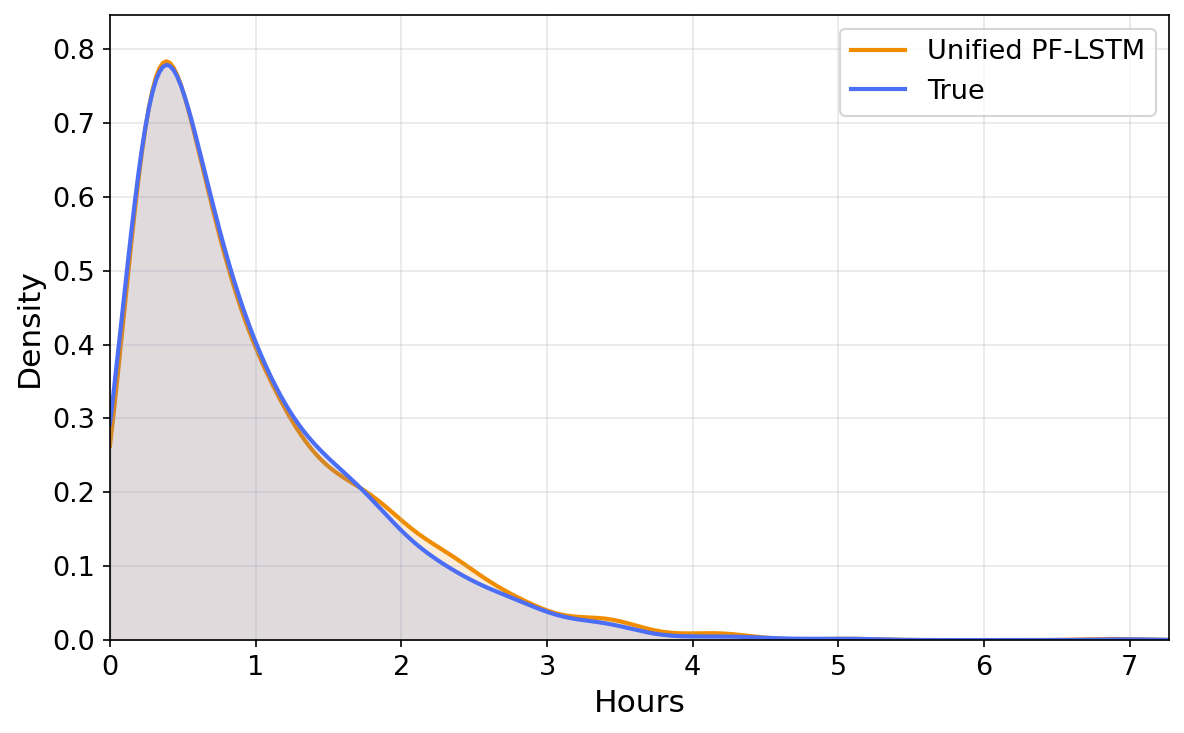}
        \caption{TPIA: ED A.}
        \label{fig:tpia_ed_a}
    \end{subfigure}
    \hfill
    \begin{subfigure}[t]{0.3\textwidth}
        \centering
        \includegraphics[width=\linewidth]
        {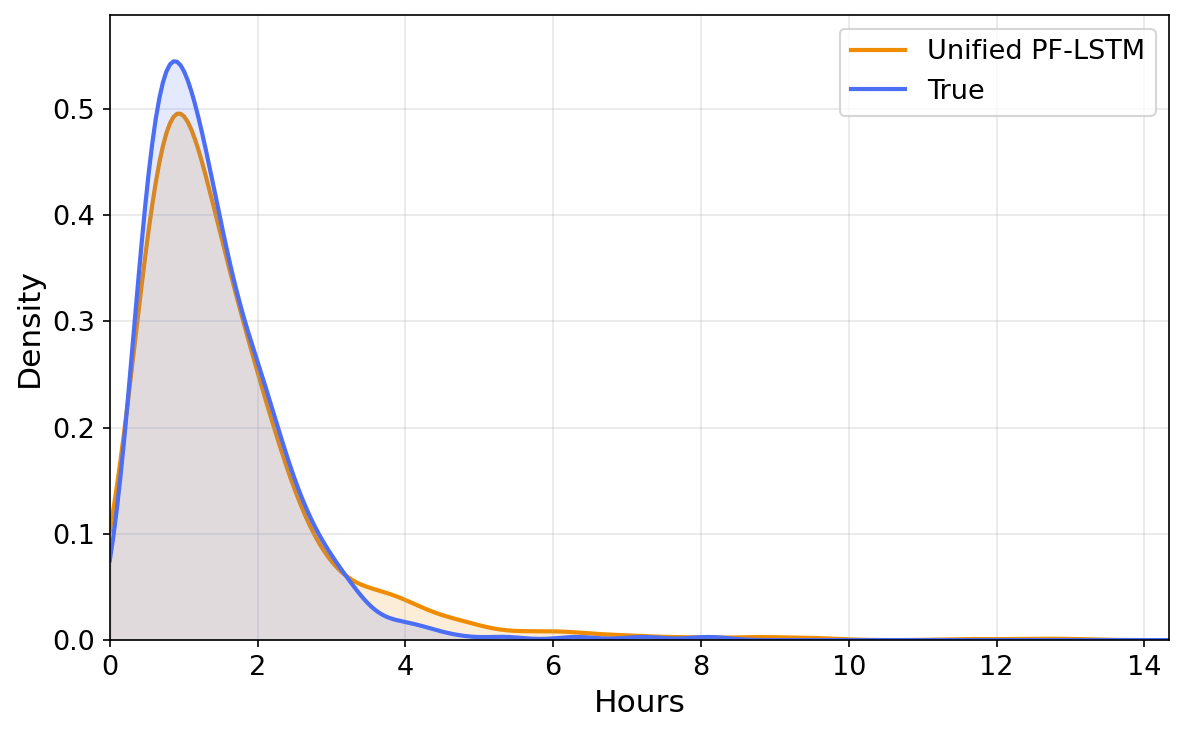}
        \caption{TPIA: ED B.}
        \label{fig:tpia_ed_b}
    \end{subfigure}
    \hfill
    \begin{subfigure}[t]{0.3\textwidth}
        \centering
        \includegraphics[width=\linewidth]
        {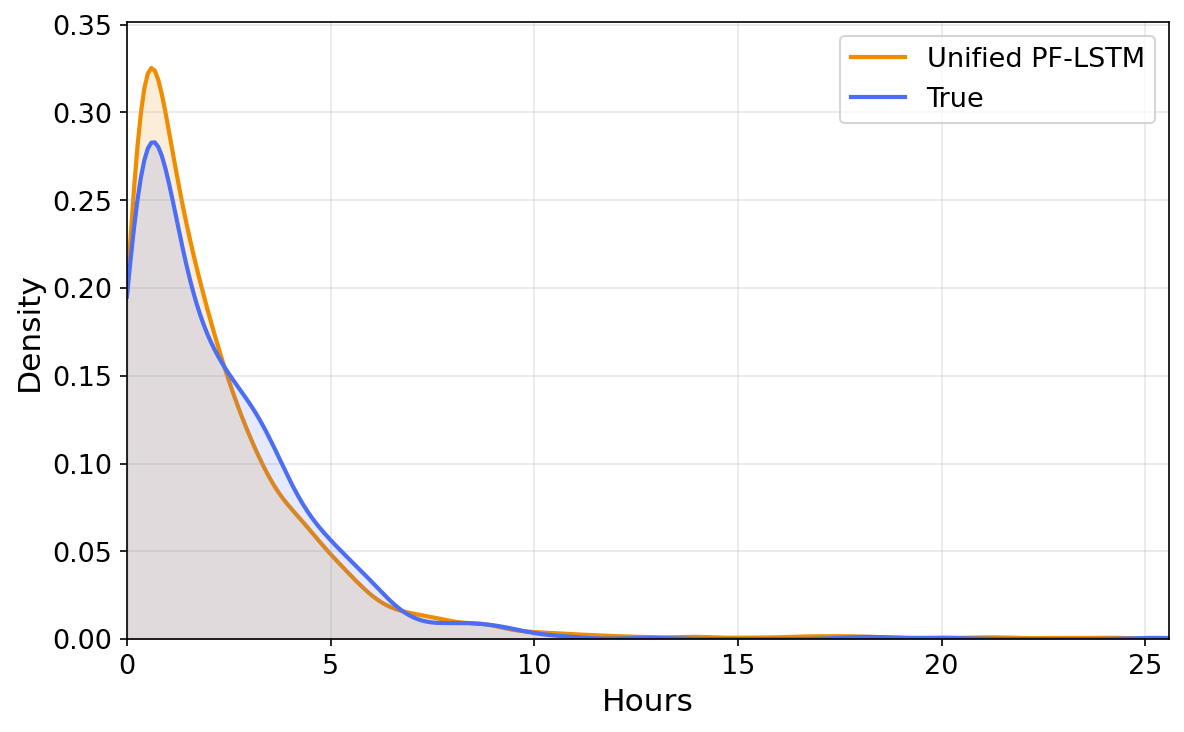}
        \caption{TPIA: ED C.}
        \label{fig:tpia_ed_c}
    \end{subfigure}

    \vspace{-0.01in}

    \begin{subfigure}[t]{0.32\textwidth}
        \centering
        \includegraphics[width=\linewidth]
        {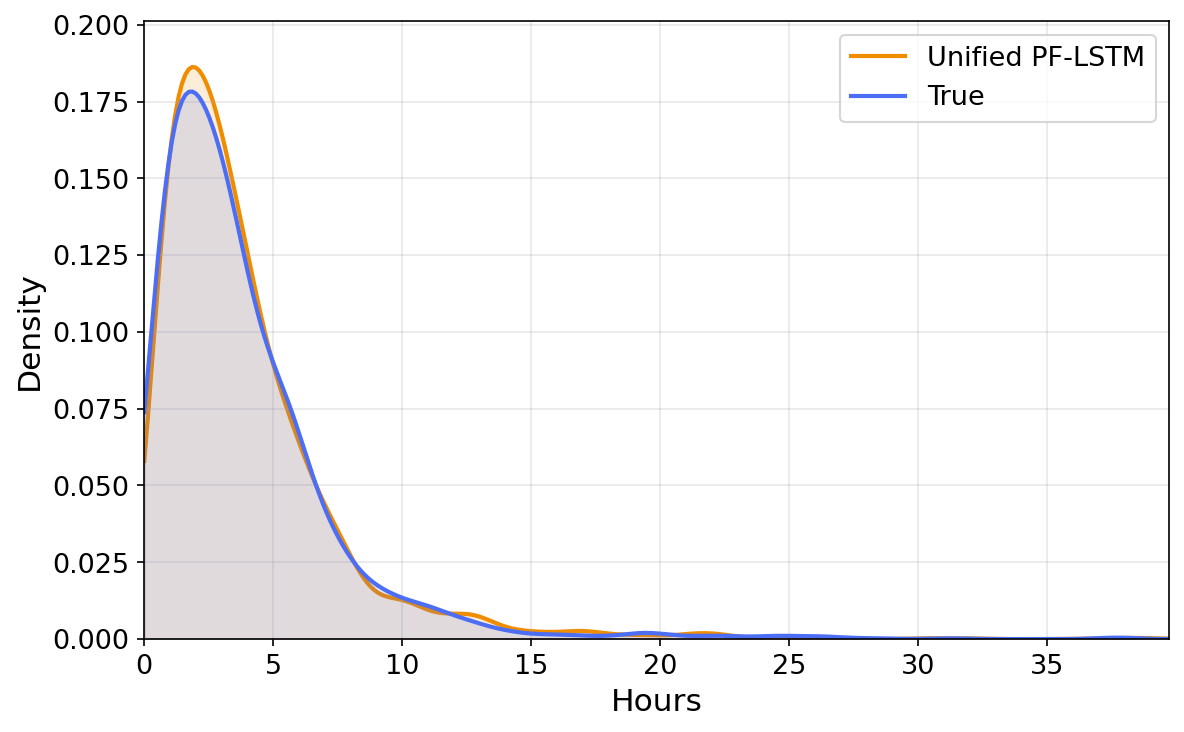}
        \caption{LOS: ED A.}
        \label{fig:los_ed_a}
    \end{subfigure}
    \hfill
    \begin{subfigure}[t]{0.32\textwidth}
        \centering
        \includegraphics[width=\linewidth]
        {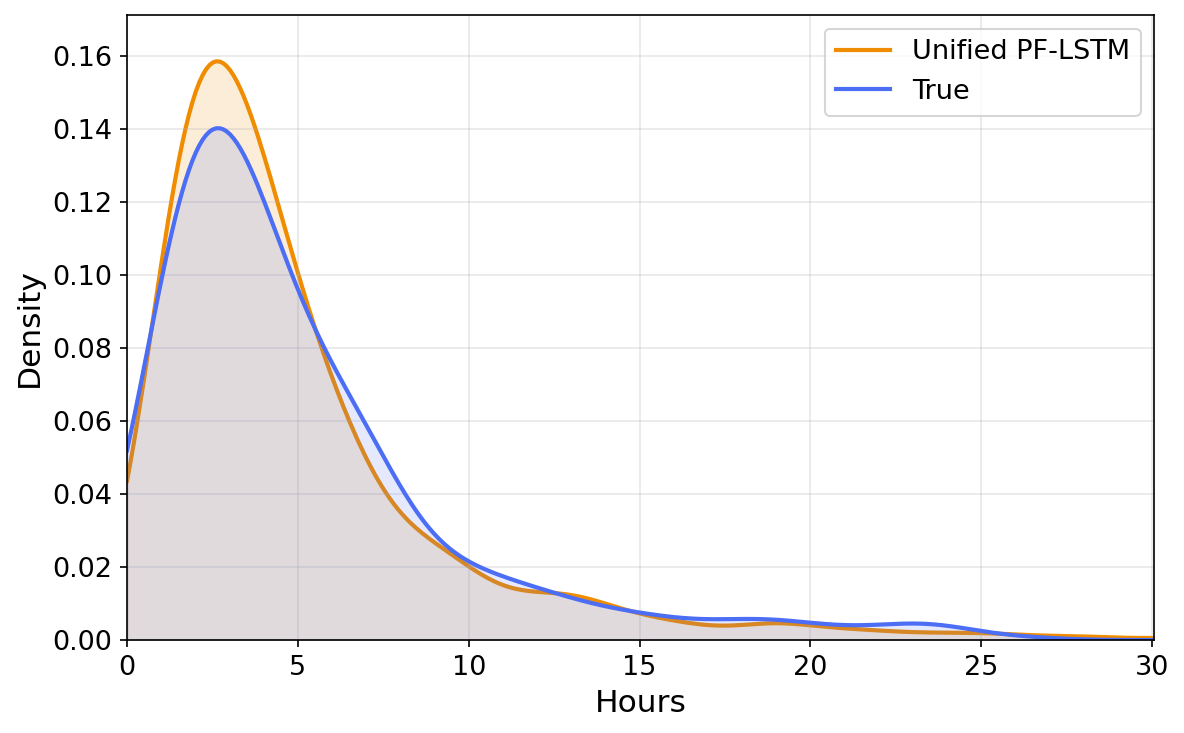}
        \caption{LOS: ED B.}
        \label{fig:los_ed_b}
    \end{subfigure}
    \hfill
    \begin{subfigure}[t]{0.32\textwidth}
        \centering
        \includegraphics[width=\linewidth]
        {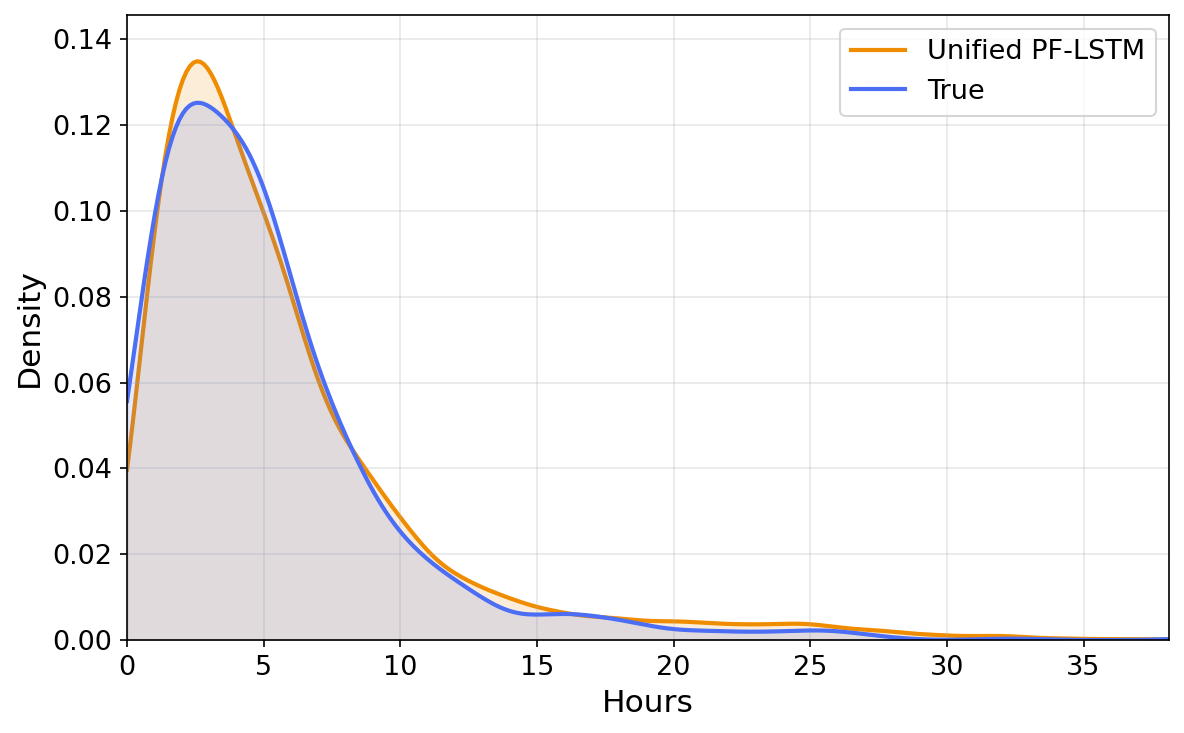}
        \caption{LOS: ED C.}
        \label{fig:los_ed_c}
    \end{subfigure}

    \vspace{-0.01in}
    \caption{Comparison of the observed (true) and Unified PF-LSTM-generated
    distributions across the three ED datasets.}
    \label{fig:tpia_los_distributions}
\end{figure}

 \vspace{0.1in}
\noindent\textbf{Results.}
Figure~\ref{fig:tpia_los_distributions} compares the observed TPIA and LOS
distributions with those generated by the proposed Unified PF-LSTM. Across all
three datasets, the generated distributions closely follow the observed
distributions.

Table~\ref{tab:model_comparison} reports the absolute percentage errors (APEs) of the mean and 90th percentile, where lower values indicate closer agreement with the held-out test data. The LSTM improves upon the RSF timing + 5-Markov routing framework, demonstrating the benefit of modeling routing and timing jointly through a sequential representation.

The Unified PF-LSTM reduces both mean and 90th-percentile errors across
all datasets and performance measures, with larger improvements observed for EDs B and C, which have more complex process structures.

TPIA errors are generally larger than LOS errors. TPIA is sensitive to early
routing errors, such as generating an incorrect activity, adding an unnecessary
activity before initial assessment, or omitting the correct assessment
destination. In contrast, errors in individual activity durations may partially
offset one another when aggregated into total LOS.

The particle-based model requires more computation than the two baselines
because each observation is propagated through multiple particles. The
reported generation times were measured on a laptop CPU. Because particle
updates can be vectorized and evaluated in parallel, a GPU implementation is expected to
reduce the wall-clock time significantly.

\begin{table}[h]
\centering
\captionsetup[subfigure]{skip=1pt}
\captionsetup{skip=2pt}
\caption{Comparison of simulation errors and generation times across the three
ED datasets. Error entries report mean APE / 90th-percentile APE. Generation
times report the total time required to generate trajectories for the
corresponding test set.}
\label{tab:model_comparison}
\resizebox{\textwidth}{!}{
\begin{tabular}{llccc}
\toprule
Metric
& Dataset
& RSF Timing + 5-Markov Routing
& LSTM
& Unified PF-LSTM \\
\midrule

\multirow{3}{*}{TPIA error (\%)}
& ED A & $17.07/32.75$ & $10.54/28.36$
& ${5.25/5.74}$ \\
& ED B & $50.00/70.48$ & $38.80/42.28$
& ${14.36/18.46}$ \\
& ED C & $27.75/28.05$ & $15.38/17.27$
& ${10.87/5.00}$ \\
\midrule

\multirow{3}{*}{LOS error (\%)}
& ED A & $12.26/11.08$ & $9.41/6.11$
& ${3.31/1.82}$ \\
& ED B & $40.33/48.91$ & $12.70/29.07$
& ${6.53/16.74}$ \\
& ED C & $23.77/36.00$ & $8.55/13.42$
& ${4.16/3.89}$ \\
\midrule

\multirow{3}{*}{Generation time (seconds)}
& ED A & $3.85$ & $77.63$ & $194.08$ \\
& ED B & $2.00$ & $150.00$ & $375.00$ \\
& ED C & $38.07$ & $202.65$ & $506.63$ \\
\bottomrule
\end{tabular}
}
\vspace{-0.15in}
\end{table}

\vspace{-0.2in}
\subsection{Sensitivity and Ablation Study}
\label{sec:sensitivity}
\vspace{-0.1in}

We examine the effects of the number of particles, the MGF-based belief
features, and the number of predicted quantiles. Due to space limitations, we
omit the detailed results and summarize the main findings below.

First, increasing the number of particles generally improves performance from
$P=1$ to $P=30$. Larger particle sets provide a richer belief representation,
although they also increase computational cost. We therefore use $P=20$ to balance predictive performance and efficiency.

Second, we compare the full Unified PF-LSTM with a mean-only variant that
excludes the MGF features. The mean-only variant performs worse, particularly
for ED B, whose event log provides less information about the underlying
system state. Among models using MGF features, increasing $M$ improves
performance by capturing distributional information beyond the mean, such as
dispersion and multimodality. However, the improvements become minor beyond
$M=8$, so we use $M=8$ in the default configuration.

Finally, increasing the number of predicted quantiles from $N_q=10$ to
$N_q=50$ improves the representation of the conditional sojourn-time
distribution. However, increasing $N_q$ from 20 to 50 provides only minor
additional improvements while noticeably increasing computation time. We
therefore use $N_q=20$ in the default configuration.

\vspace{-.2in}
\section{Conclusion and Future Directions}
\vspace{-.1in}

We introduced a Unified PF-LSTM framework for data-driven process simulation.  The framework
maintains a particle-based belief over recurrent-state hypotheses that encode
plausible latent process conditions and uses MGF-based features to retain
information about the shape of this belief. The resulting representation
supports the generation of complete case trajectories while representing
latent-state uncertainty induced by partial observability of operational
conditions, as well as variability in routing and sojourn-time outcomes.
\\
Our experimental results using three real-world ED datasets with different
levels of process complexity and information availability show that our 
framework outperforms the alternative data-driven baselines in reproducing
routing behavior, activity durations, and system-level performance measures,
with particularly strong improvements in more complex settings.

Future work may extend the framework to other service systems and investigate
online adaptation under changing operating conditions. Additional directions
include using the learned model for counterfactual analysis and integrating it
with reinforcement learning for policy evaluation, optimization, and
operational decision-making.

\bibliographystyle{plainnat}
\bibliography{refs}


\end{document}